\documentclass{article}

\newcommand{\unorm}[1]{\|#1\|}
\newcommand{\unorms}[1]{\unorm{#1}^2}

\newcommand{\indep}{\mathop{\perp\!\!\!\perp}}

\newcommand{\mathbbE}{\mathbb{E}}

\newcommand{\mathbbR}{\mathbb{R}}
\newcommand{\mathbbS}{\mathbb{S}}

\newcommand{\boldzero}{{\boldsymbol{0}}}
\newcommand{\boldone}{{\boldsymbol{1}}}

\newcommand{\boldA}{{\boldsymbol{A}}}

\newcommand{\boldD}{{\boldsymbol{D}}}

\newcommand{\boldH}{{\boldsymbol{H}}}
\newcommand{\boldI}{{\boldsymbol{I}}}

\newcommand{\boldK}{{\boldsymbol{K}}}
\newcommand{\boldL}{{\boldsymbol{L}}}

\newcommand{\boldR}{{\boldsymbol{R}}}

\newcommand{\boldW}{{\boldsymbol{W}}}

\newcommand{\boldh}{{\boldsymbol{h}}}

\newcommand{\boldw}{{\boldsymbol{w}}}
\newcommand{\boldx}{{\boldsymbol{x}}}
\newcommand{\boldy}{{\boldsymbol{y}}}
\newcommand{\boldz}{{\boldsymbol{z}}}

\newcommand{\boldalpha}{{\boldsymbol{\alpha}}}
\newcommand{\boldbeta}{{\boldsymbol{\beta}}}

\newcommand{\boldmu}{{\boldsymbol{\mu}}}

\newcommand{\boldGamma}{{\boldsymbol{\Gamma}}}

\newcommand{\boldSigma}{{\boldsymbol{\Sigma}}}

\newcommand{\calS}{{\mathcal{S}}}

\newcommand{\calX}{{\mathcal{X}}}
\newcommand{\calY}{{\mathcal{Y}}}
\newcommand{\calZ}{{\mathcal{Z}}}

\newcommand{\boldbetah}{{\widehat{\boldbeta}}}

\newcommand{\pxy}{p_{\mathrm{xy}}}
\newcommand{\pzy}{p_{\mathrm{zy}}}

\newcommand{\pz}{p_{\mathrm{z}}}
\newcommand{\py}{p_{\mathrm{y}}}

\newcommand{\boldHh}{{\widehat{\boldH}}}
\newcommand{\boldhh}{{\widehat{\boldh}}}

\newcommand{\boldalphah}{\widehat{\boldalpha}}

\usepackage{graphicx} 
\usepackage{subfigure} 

\usepackage{natbib}

\usepackage{algorithm}
\usepackage{algorithmic}

\usepackage{hyperref}

\usepackage{subfigure,amsmath,amssymb,bm} 

\newcommand{\LSDRG}{\textnormal{LSDR}}

\usepackage[accepted]{icml2011}

\icmltitlerunning{Sufficient Component Analysis}

\begin{document} 

\twocolumn[
\icmltitle{Sufficient Component Analysis for Supervised Dimension Reduction}

\icmlauthor{Makoto Yamada}{yamada@sg.cs.titech.ac.jp}
\icmlauthor{Gang Niu}{gang@sg.cs.titech.ac.jp}
\icmlauthor{Jun Takagi}{takagi@sg.cs.titech.ac.jp}
\icmlauthor{Masashi Sugiyama}{sugi@cs.titech.ac.jp}
\icmladdress{Tokyo Institute of Technology, Tokyo 152-8552, Japan}

\icmlkeywords{boring formatting information, machine learning, ICML}

\vskip 0.3in
]

\begin{abstract} 
The purpose of sufficient dimension reduction (SDR) is to find the
low-dimensional subspace of input features that is sufficient for
predicting output values. In this paper, we propose a novel \emph{distribution-free}
 SDR method called \emph{sufficient component analysis} (SCA),
which is computationally more efficient than existing methods.
In our method, a solution is computed
by iteratively performing dependence estimation and maximization:
Dependence estimation is analytically carried out
by recently-proposed \emph{least-squares mutual information} (LSMI),
and dependence maximization is also analytically carried out
by utilizing the \emph{Epanechnikov kernel}.
Through large-scale experiments on real-world image classification  and audio tagging problems,
the proposed method is shown to compare favorably with
existing dimension reduction approaches.
\end{abstract}

\section{Introduction}
The goal of \emph{sufficient dimension reduction} (SDR) is to learn a
transformation matrix $\boldW$ from input feature $\boldx$ to its low-dimensional
representation $\boldz$ ($=\boldW\boldx$) which has `sufficient' information for predicting
output value $\boldy$.  SDR can be formulated as the problem of finding
$\boldz$ such that $\boldx$ and $\boldy$ are conditionally independent
given $\boldz$
\cite{book:Cook:1998,TAS:Fukumizu+etal:2009}.

Earlier SDR methods developed in statistics community, such as
\emph{sliced inverse regression} \cite{JASA:Li:1991},
\emph{principal Hessian direction} \cite{JASA:Li:1992},
and 
\emph{sliced average variance estimation} \cite{TAM:Cook:2000},
rely on the elliptical assumption (e.g., Gaussian) of the data,
which may not be fulfilled in practice.

To overcome the limitations of these approaches,
the \emph{kernel dimension reduction} (KDR) was proposed \cite{TAS:Fukumizu+etal:2009}.
KDR employs a kernel-based dependence measure,
which does not require the elliptical assumption (i.e., distribution-free),
and the solution $\boldW$ is computed by a gradient method.
Although KDR is a highly flexible SDR method,
its critical weakness is the kernel function choice---the
performance of KDR depends on the choice of kernel functions
and the regularization parameter,
but there is no systematic model selection method available.
Furthermore, KDR scales poorly to massive datasets
since the gradient-based optimization is computationally demanding.
Another important limitation of KDR in practice is that
there is no good way to set an initial solution---many
random restarts may be needed for finding a good local optima,
which makes the entire procedure even slower
and the performance of dimension reduction unstable.


To overcome the limitations of KDR,
a novel SDR method called \emph{least-squares dimension reduction} (LSDR) was proposed recently 
\cite{AISTATS:Suzuki+Sugiyama:2010}. 
LSDR adopts a squared-loss variant of mutual information
as a dependency measure, which is efficiently estimated by \emph{least-squares mutual information}
(LSMI) \cite{BMCBio:Suzuki+etal:2009a}.  A notable advantage of LSDR over KDR
is that kernel functions and its tuning parameters such as the kernel width and the regularization parameter
can be naturally optimized by cross-validation. 
However, LSDR still relies on a computationally expensive gradient method
and there is no good initialization scheme.

In this paper, we  propose a novel SDR method called \emph{sufficient component analysis} (SCA),
which can overcome the computational inefficiency of LSDR. 
In SCA,
the solution $\boldW$ in each iteration is obtained \emph{analytically} 
by just solving an eigenvalue problem,
which significantly contributes to improving the computational efficiency.
Moreover, based on the above analytic-form solution,
we develop a method to design a good initial value for optimization,
which further reduces the computational cost and help obtain a good local optimum solution.

Through large-scale experiments using
the \emph{PASCAL Visual Object Classes (VOC) 2010} dataset \cite{pascal-voc-2010} 
and the \emph{Freesound} dataset \cite{ref:freesound},
we demonstrate the usefulness of the proposed method.


\section{Sufficient Dimension Reduction with Squared-Loss Mutual Information}
In this section, we formulate the problem of \emph{sufficient dimension reduction} (SDR)
based on \emph{squared-loss mutual information} (SMI).


\subsection{Problem Formulation}
Let $\calX (\subset \mathbbR^d)$ be the domain of input feature $\boldx$
and $\calY$ be the domain of output data\footnote{
$\calY$ could be either continuous (i.e., regression) or categorical (i.e., classification).
Multi-dimensional outputs (e.g., multi-task regression and multi-label classification)
and structured outputs (such as sequences, trees, and graphs) can also be handled
in the proposed framework.
} $\boldy$.
Suppose we are given $n$ independent and identically distributed (i.i.d.) paired samples,
\[
D^n = \{(\boldx_i, \boldy_i)~|~\boldx_i \in \calX, ~\boldy_i \in \calY,~i=1,\ldots,n\},
\]
drawn from a joint distribution with density $\pxy(\boldx, \boldy)$.

The goal of SDR is to find a low-dimensional representation $\boldz$ 
($\in \mathbbR^m$, $m\le d$) of input $\boldx$
that is sufficient to describe output $\boldy$.
More precisely, we find $\boldz$ such that
\begin{align}
\boldy \indep \boldx~|~\boldz,
\label{eq:conditional_indep}
\end{align}
meaning that, given the projected feature $\boldz$,
the feature $\boldx$ is conditionally independent of output $\boldy$. 

In this paper, we focus on linear dimension reduction scenarios:
\[
\boldz = \boldW \boldx,
\]
where $\boldW$ is a transformation matrix.
$\boldW$ belongs to the \emph{Stiefel manifold} $\mathbbS_{m}^d(\mathbbR)$:
\[
\mathbbS_{m}^d(\mathbbR) := \{\boldW \in \mathbbR^{m \times d} | \boldW \boldW^\top = \boldI_m\},
\]
where $^\top$ denotes the transpose
and $\boldI_m$ is the $m$-dimensional identity matrix.
Below, we assume that the reduced dimension $m$ is known.


\subsection{Dependence Estimation-Maximization Framework}
\label{subsec:DEM}
\citet{AISTATS:Suzuki+Sugiyama:2010} showed that
the optimal transformation matrix that leads to Eq.\eqref{eq:conditional_indep}
can be characterized as
\begin{align}
\boldW^\ast &= \mathop{\textnormal{argmax}}_{\boldW\in\mathbbR^{m \times d}}~ \textnormal{SMI}(Z, Y) 
~~\text{s.t.}~\boldW\boldW^\top = \boldI_m.
\label{eq:argmaxSMI}
\end{align}
In the above, ${\mathrm{SMI}}(Z,Y)$ 
is the \emph{squared-loss mutual information}:
\begin{align*}
{\mathrm{SMI}}(Z,Y)
&:= \frac{1}{2}\mathbbE_{\pz,\py}\left[
\left( \frac{\pzy(\boldz,\boldy)}{\py(\boldy)\pz(\boldz)}  - 1\right)^2
\right],
\end{align*}
where $\mathbbE_{\pz,\py}$
denotes the expectation over
the marginals $\pz(\boldz)$ and $\py(\boldy)$.
Note that SMI is the \emph{Pearson divergence}
from $\pzy(\boldz, \boldy)$ to $\pz(\boldz)\py(\boldy)$,
while the ordinary mutual information is the Kullback-Leibler divergence
from $\pzy(\boldz, \boldy)$ to $\pz(\boldz)\py(\boldy)$.
The Pearson divergence and the Kullback-Leibler divergence both 
belong to the class of $f$-divergences,
which shares similar theoretical properties.
For example, SMI is non-negative and is zero if and only if
$Z$ and $Y$ are statistically independent,
as ordinary mutual information.

Based on Eq.\eqref{eq:argmaxSMI},
we develop the following iterative algorithm for learning $\boldW$:
\begin{description}
  \item[(i) Initialization:] Initialize the transformation matrix $\boldW$
    (see Section~\ref{subsec:initialization}).
  \item[(ii) Dependence estimation:] For current $\boldW$,
    an SMI estimator $\widehat{\textnormal{SMI}}$ is obtained
    (see Section~\ref{subsec:dependence-estimation}).
  \item[(iii) Dependence maximization:] Given an SMI estimator $\widehat{\textnormal{SMI}}$,
    its maximizer with respect to $\boldW$ is obtained
    (see Section~\ref{subsec:dependence-maximization}).
  \item[(iv) Convergence check:] The above (ii) and (iii) are repeated
    until $\boldW$ fulfills some convergence criterion\footnote{
      In experiments, we used the criterion that 
      the improvement of $\widehat{\textnormal{SMI}}$ is less than $10^{-6}$.
    }.
\end{description}


\section{Proposed Method: Sufficient Component Analysis}
In this section, we describe our proposed method
called the \emph{sufficient component analysis} (SCA).

\subsection{Dependence Estimation}
\label{subsec:dependence-estimation}
In SCA, we utilize a non-parametric SMI estimator called
\emph{least-squares mutual information} (LSMI) \cite{BMCBio:Suzuki+etal:2009a},
which was shown to achieve the optimal convergence rate
\cite{AISTATS:Suzuki+Sugiyama:2010}.
Here, we review LSMI.


\subsubsection{Basic Idea}
A key idea of LSMI is to directly estimate the \emph{density ratio},
\[
w(\boldz, \boldy) = \frac{\pzy(\boldz, \boldy)}{\pz(\boldz)\py(\boldy)},
\]
without going through density estimation of $\pzy(\boldz, \boldy)$, 
$\pz(\boldz)$, and $\py(\boldy)$.
Here, the density ratio function $w(\boldz, \boldy)$ is
directly modeled by 
\begin{eqnarray}
w_{\boldalpha}(\boldz, \boldy)= \sum_{\ell = 1}^n \alpha_{\ell} K(\boldz,\boldz_\ell)L(\boldy,\boldy_\ell),
\label{ratio-model}
\end{eqnarray}
where $K(\boldz,\boldz')$ and $L(\boldy,\boldy')$ are kernel functions for $\boldz$
and $\boldy$, respectively.

Then, the parameter $\boldalpha=(\alpha_1,\ldots,\alpha_n)^\top$ is learned so that the following 
squared error is minimized:
\begin{align*}
J_0(\boldalpha)=\frac{1}{2}
\mathbbE_{\pz,\py}\left[
(w_{\boldalpha}(\boldz,\boldy) - w(\boldz,\boldy))^2
\right].
\end{align*}
$J_0$ can be expressed as
\begin{align*}
J_0(\boldalpha) = J(\boldalpha)+\textnormal{SMI}(Z,Y)+\frac{1}{2},
\end{align*}
where
\begin{align*}
 J(\boldalpha) &=
\frac{1}{2}\boldalpha^\top \boldH \boldalpha -\boldh^\top\boldalpha, \\
H_{\ell,\ell'} &= 
\mathbbE_{\pz,\py}\left[
K(\boldz,\boldz_\ell)L(\boldy,\boldy_\ell)
K(\boldz,\boldz_{\ell'})L(\boldy,\boldy_{\ell'})
\right],\\
h_{\ell} &= 
\mathbbE_{\pzy}\left[
K(\boldz,\boldz_\ell)L(\boldy,\boldy_\ell)
\right],
\end{align*}
and $\textnormal{SMI}(Z,Y)$ is constant with respect to $\boldalpha$. Thus, minimizing $J_0$ is equivalent to minimizing $J$.

\subsubsection{Computing the Solution}

Approximating the expectations in $\boldH$ and $\boldh$ included in $J$
by empirical averages,
we arrive at the following optimization problem:
\begin{align*}
\min_{\boldalpha} \left[
  \frac{1}{2}\boldalpha^\top \widehat{\boldH}\boldalpha
  -\widehat{\boldh}^\top\boldalpha
 + \lambda \boldalpha^\top \boldR \boldalpha \right], 
\end{align*}
where a regularization term $\lambda\boldalpha^\top \boldR \boldalpha$ is included
for avoiding overfitting, $\lambda$ ($\ge0$) is a regularization parameter, $\boldR$ is a regularization matrix, and,
for $\boldz_i = \boldW\boldx_i$,
\begin{align*}
\widehat{H}_{\ell,\ell'} &= \frac{1}{n^2} 
\sum_{i,j = 1}^nK(\boldz_i,\boldz_\ell)L(\boldy_i,\boldy_\ell)
K(\boldz_j,\boldz_{\ell'})L(\boldy_j,\boldy_{\ell'}),\\
\widehat{h}_\ell &= \frac{1}{n}\sum_{i = 1}^n 
K(\boldz_i,\boldz_\ell)L(\boldy_i,\boldy_\ell).
\end{align*}

Differentiating the above objective function with respect to $\boldalpha$ and equating it to zero,
we can obtain an analytic-form solution:
\begin{align}
\label{eq:LSMI_Solution}
\widehat{\boldalpha}= (\widehat{\boldH} + \lambda \boldR)^{-1}\widehat{\boldh}.
\end{align}

Based on the fact that ${\mathrm{SMI}}(Z,Y)$ is expressed as 
\begin{align*}
{\mathrm{SMI}}(Z,Y)
=\frac{1}{2} 
\mathbbE_{\pzy}\left[
 w(\boldz,\boldy)\right]-\frac{1}{2},
\end{align*}
the following SMI estimator can be obtained:
\begin{align}
\widehat{\mathrm{SMI}}&= 
\frac{1}{2}\boldhh^\top\boldalphah -\frac{1}{2}. 
\label{SMIhat}
\end{align}


\subsubsection{Model Selection}
\label{subsubsec:LSMI-CV}
Hyper-parameters included in the kernel functions
and the regularization parameter can be optimized by cross-validation 
with respect to $J$.

More specifically, the samples $\calZ=\{(\boldz_i, \boldy_i)\}_{i=1}^n$ are
divided into $K$ disjoint subsets $\{\calZ_k\}_{k=1}^K$ of (approximately) the same size.
Then, an estimator $\widehat{\boldalpha}_{\calZ_k}$ is obtained using $\calZ\backslash\calZ_k$
(i.e,. all samples without $\calZ_k$),
and the approximation error for the hold-out samples $\calZ_k$ is computed as 
\[
J_{\calZ_k}^{(K \text{-CV})} = \frac{1}{2}\widehat{\boldalpha}_{\calZ_k}^\top \widehat{\boldH}_{\calZ_k}\widehat{\boldalpha}_{\calZ_k} - \widehat{\boldh}_{\calZ_k}^\top\widehat{\boldalpha}_{\calZ_k},
\]
where,
for $|\calZ_k|$ being the number of samples in the subset $\calZ_k$, 
\begin{align*}
[\widehat{H}_{\calZ_k}]_{\ell,\ell'} &= \frac{1}{|\calZ_k|^2} 
\sum_{(\boldz,\boldy),(\boldz',\boldy')\in\calZ_k}
K(\boldz,\boldz_\ell)L(\boldy,\boldy_\ell)\\
&~~~~~~~~~~~~~\times K(\boldz',\boldz_{\ell'})L(\boldy',\boldy_{\ell'}),\\
[\widehat{h}_{\calZ_k}]_\ell &= \frac{1}{|\calZ_k|}\sum_{(\boldz,\boldy)\in\calZ_k}
K(\boldz,\boldz_\ell)L(\boldy,\boldy_\ell).
\end{align*}

This procedure is repeated for $k = 1, \ldots, K$,
and its average $J^{(K\text{-CV})}$ is outputted as
\begin{align*}
  J^{(K\text{-CV})} = \frac{1}{K}\sum_{k = 1}^K J_{\calZ_k}^{(K\text{-CV})}.
\end{align*}
We compute $J^{(K\text{-CV})}$ for all model candidates, and choose the model that minimizes $J^{(K\text{-CV})}$. 

\subsection{Dependence Maximization}
\label{subsec:dependence-maximization}
Given an SMI estimator $\widehat{\mathrm{SMI}}$ \eqref{SMIhat},
we next show how $\widehat{\mathrm{SMI}}$ can be efficiently maximized
with respect to $\boldW$:
\begin{align*}
\max_{\boldW \in \mathbbR^{m \times d}}\widehat{\mathrm{SMI}}
~~~\textnormal{s.t.}~\boldW\boldW^\top = \boldI_m.
\end{align*}
We propose to use a truncated negative quadratic function
called the \emph{Epanechnikov kernel} \cite{TPA:Epanechnikov:1969}
as a kernel for $\boldz$:
\begin{align*}
K(\boldz,\boldz_\ell) &= \max\left(0,1 - \frac{\| \boldz - \boldz_\ell \|^2}{2 \sigma_{\mathrm z}^2}\right).
\end{align*}

Let $I(c)$ be the indicator function, i.e., $I(c)=1$ if $c$ is true and zero otherwise.
Then, for the above kernel, $\widehat{\mathrm{SMI}}$ can be expressed as
\begin{align*}
\widehat{\mathrm{SMI}}=
\frac{1}{2}{\textnormal{tr}}\left(\boldW \boldD \boldW^\top\right) - \frac{1}{2}, 
\end{align*}
where $\textnormal{tr}(\boldA)$ is the trace of matrix $\boldA$, and
\begin{align*}
\boldD &=  \frac{1}{n}\sum_{i=1}^{n} \sum_{\ell =1}^n \widehat{\alpha}_{\ell}(\boldW)
I\left(\frac{\unorms{\boldW\boldx_i- \boldW\boldx_\ell}}{2\sigma_{\mathrm z}^2} < 1\right)\\
&~~~\times L(\boldy_i,\boldy_\ell)
 \left[\frac{1}{m}\boldI_d -  \frac{1}{2\sigma_{\mathrm z}^2} (\boldx_i -\boldx_\ell)(\boldx_i - \boldx_\ell)^\top \right]. 
\end{align*}
Here, by $\widehat{\alpha}_{\ell}(\boldW)$, we explicitly indicated the fact
that $\widehat{\alpha}_{\ell}$ depends on $\boldW$.

Let $\boldD'$ be $\boldD$ with $\boldW$ replaced by $\boldW'$,
where $\boldW'$ is a transformation matrix obtained in the previous iteration.
Thus, $\boldD'$ no longer depends on $\boldW$.
Here we replace $\boldD$ in $\widehat{\mathrm{SMI}}$
by $\boldD'$, which gives the following simplified SMI estimate:
\begin{align}
\frac{1}{2}{\textnormal{tr}}\left(\boldW \boldD' \boldW^\top\right) - \frac{1}{2}.
\label{SMIhat-simplified}
\end{align}
A maximizer of Eq.\eqref{SMIhat-simplified}
can be analytically obtained by
$(\boldw_1|\cdots|\boldw_m)^\top$,
where $\{\boldw_i\}_{i=1}^m$ are the $m$ principal components of $\boldD'$.

\subsection{Initialization of $\boldW$}
\label{subsec:initialization}
In the dependence estimation-maximization framework
described in Section~\ref{subsec:DEM},
initialization of the transformation matrix $\boldW$
is important. 
Here we propose to initialize it
based on dependence maximization without dimensionality reduction.

More specifically, we determine the initial transformation matrix
as $(\boldw_1^{(0)}|\cdots|\boldw_m^{(0)})^\top$,
where $\{\boldw_i^{(0)}\}_{i=1}^m$ are the $m$ principal components of
$\boldD^{(0)}$:
\begin{align*}
\boldD^{(0)} &=  \frac{1}{n}\sum_{i=1}^{n} \sum_{\ell =1}^n \widehat{\alpha}_{\ell}^{(0)}
I\left(\frac{\unorms{\boldx_i- \boldx_\ell}}{2\sigma_{\mathrm x}^2} < 1\right)
L(\boldy_i,\boldy_\ell)\\
&~~~\times \left[\frac{1}{m}\boldI_d -  \frac{1}{2\sigma_{\mathrm x}^2} (\boldx_i -\boldx_\ell)(\boldx_i - \boldx_\ell)^\top \right], \\
\widehat{\boldalpha}^{(0)} &=
(\widehat{\boldH}^{(0)} + \lambda \boldR)^{-1}\widehat{\boldh}^{(0)},\\
\widehat{H}^{(0)}_{\ell,\ell'} &= \frac{1}{n^2} 
\sum_{i,j = 1}^nK'(\boldx_i,\boldx_\ell)L(\boldy_i,\boldy_\ell)\\
&~~~~~~~\times K'(\boldx_j,\boldx_{\ell'})L(\boldy_j,\boldy_{\ell'}),\\
\widehat{h}^{(0)}_\ell &= \frac{1}{n}\sum_{i = 1}^n 
K'(\boldx_i,\boldx_\ell)L(\boldy_i,\boldy_\ell),\\
K'(\boldx,\boldx_\ell) &= \max\left(0,1 - \frac{\| \boldx - \boldx_\ell \|^2}{2 \sigma_{\mathrm x}^2}\right).
\end{align*}
$\sigma_{{\mathrm x}}$ is the kernel width
and is chosen by cross-validation (see Section~\ref{subsubsec:LSMI-CV}). 


\section{Relation to Existing Methods}
Here, we review existing SDR methods and discuss the relation to the proposed SCA method.

\subsection{Kernel Dimension Reduction}
\emph{Kernel Dimension Reduction} (KDR) \cite{TAS:Fukumizu+etal:2009} 
tries to directly maximize the conditional independence of
$\boldx$ and $\boldy$ given $\boldz$
under a kernel-based independence measure.

The KDR learning criterion is given by
\begin{align}
{\boldW}^\ast &= \mathop{\textnormal{argmax}}_{\boldW \in \mathbbR^{m \times d}}
\textnormal{tr}\left[\widetilde{\boldL} (\widetilde{\boldK} + n\epsilon \boldI_n)^{-1}\right] \nonumber \\
&~~~~~\text{s.t.}~\boldW\boldW^\top = \boldI_m,
\end{align}
where  $\widetilde{\boldL} = \boldGamma \boldL \boldGamma$,
$\boldGamma = \boldI - \frac{1}{n}\boldone_n \boldone_n^\top$,
$L_{i,j} = L(\boldy_i,\boldy_j)$, 
$\widetilde{\boldK} = \boldGamma \boldK \boldGamma$,
$K_{i,j} = K(\boldz_i,\boldz_j)$, and
$\epsilon$ is a regularization parameter.

Solving the above optimization problem is cumbersome 
since the objective function is non-convex.
In the original KDR paper \cite{TAS:Fukumizu+etal:2009},
a gradient method is employed for finding a local optimal solution.
However, the gradient-based optimization is computationally demanding
due to its slow convergence and it requires many restarts for finding a good local optima.
Thus, KDR scales poorly to massive datasets.

Another critical weakness of KDR is the kernel function choice.
The performance of KDR depends on the choice of kernel functions
and the regularization parameter,
but there is no systematic model selection method for KDR available. 
Using the Gaussian kernel with its width set to the median distance
between samples is a standard heuristic in practice,
but this does not always work very well.

Furthermore, KDR lacks
a good way to set an initial solution in the gradient procedure.
Then, in practice, we need to run the algorithm many
times with random initial points for finding good local optima.
However, this makes the entire procedure even slower
and the performance of dimension reduction unstable.

The proposed SCA method can successfully overcome the above weaknesses of KDR---SCA
is equipped with cross-validation for model selection (Section~\ref{subsubsec:LSMI-CV}),
its solution can be computed analytically
(see Section~\ref{subsec:dependence-maximization}),
and a systematic initialization scheme is available
(see Section~\ref{subsec:initialization}).

\subsection{Least-Squares Dimensionality Reduction}
\emph{Least-squares dimension reduction} (LSDR) is a recently
proposed SDR method that can overcome the limitations of KDR \cite{AISTATS:Suzuki+Sugiyama:2010}.
That is, LSDR is equipped with a natural model selection procedure
based on cross-validation.

The proposed SCA can actually be regarded as a computationally efficient
alternative to LSDR.
Indeed, LSDR can also be interpreted as a dependence estimation-maximization algorithm
(see Section~\ref{subsec:DEM}),
and the dependence estimation procedure is essentially the same as the proposed SCA,
i.e., LSMI is used.
The dependence maximization procedure is different from SCA---LSDR uses
a \emph{natural gradient} method \cite{NECO:Amari:1998}.

In LSDR, the following SMI estimator is used:
\begin{align*}
\widetilde{\textnormal{SMI}}=
\boldalphah^\top \boldhh -  \frac{1}{2}\boldalphah^\top \boldHh \boldalphah - \frac{1}{2},
\end{align*}
where $\boldalphah$, $\boldhh$ and $\boldHh$ are defined in Section~\ref{subsec:dependence-estimation}. 
Then the gradient of $\widetilde{\textnormal{SMI}}$ is given by
\begin{align*}
\frac{\partial \widetilde{\textnormal{SMI}}}{\partial W_{\ell,\ell'}} &= \frac{\partial \widehat{\boldh}^\top}{\partial W_{\ell,\ell'}}(2\boldalphah - \boldbetah)
- \boldalphah^\top \frac{\partial \widehat{\boldH}}{\partial W_{\ell,\ell'}}(\frac{3}{2}\boldalphah - \boldbetah)\\
&~~~~+ \boldalphah^{\top} \frac{\partial \boldR}{\partial W_{\ell,\ell'}}(\boldbetah - \boldalphah),
\end{align*}
where $\boldbetah = (\widehat{\boldH} + \lambda \boldR)^{-1}\widehat{\boldH}\boldalphah$. The \emph{natural gradient} update of $\boldW$, 
which takes into account the structure of the Stiefel manifold
\cite{NECO:Amari:1998}, 
is given by
\begin{align*}
\boldW
\leftarrow \boldW \exp\left(\eta\Bigg(\boldW^\top
 \frac{\partial \widetilde{\textnormal{SMI}}}{\partial \boldW}
 - \frac{\partial \widetilde{\textnormal{SMI}}}{\partial \boldW}^\top \boldW\Bigg)\right), 
\end{align*}
where `$\exp$' for a matrix denotes the \emph{matrix exponential}.
$\eta \geq 0$ is a step size, which may be optimized by a line-search method such as \emph{Armijo's rule} \cite{Book:Patriksson:1999}.

Since cross-validation is available for model selection of LSMI,
LSDR is more favorable than KDR.
However, its optimization still relies on a gradient-based method
and thus it is computationally expensive.

Furthermore, there seems no good initialization scheme of
the transformation matrix $\boldW$.
In the original paper by \citet{AISTATS:Suzuki+Sugiyama:2010},
initial values were chosen randomly and the gradient method was run many times
for finding a better local solution.

The proposed SCA method can successfully overcome the above weaknesses of LSDR,
by providing an analytic-form solution (see Section~\ref{subsec:dependence-maximization})
and a systematic initialization scheme (see Section~\ref{subsec:initialization}).

\begin{figure*}[t]
  \centering
\subfigure{
\includegraphics[width=.23\textwidth]{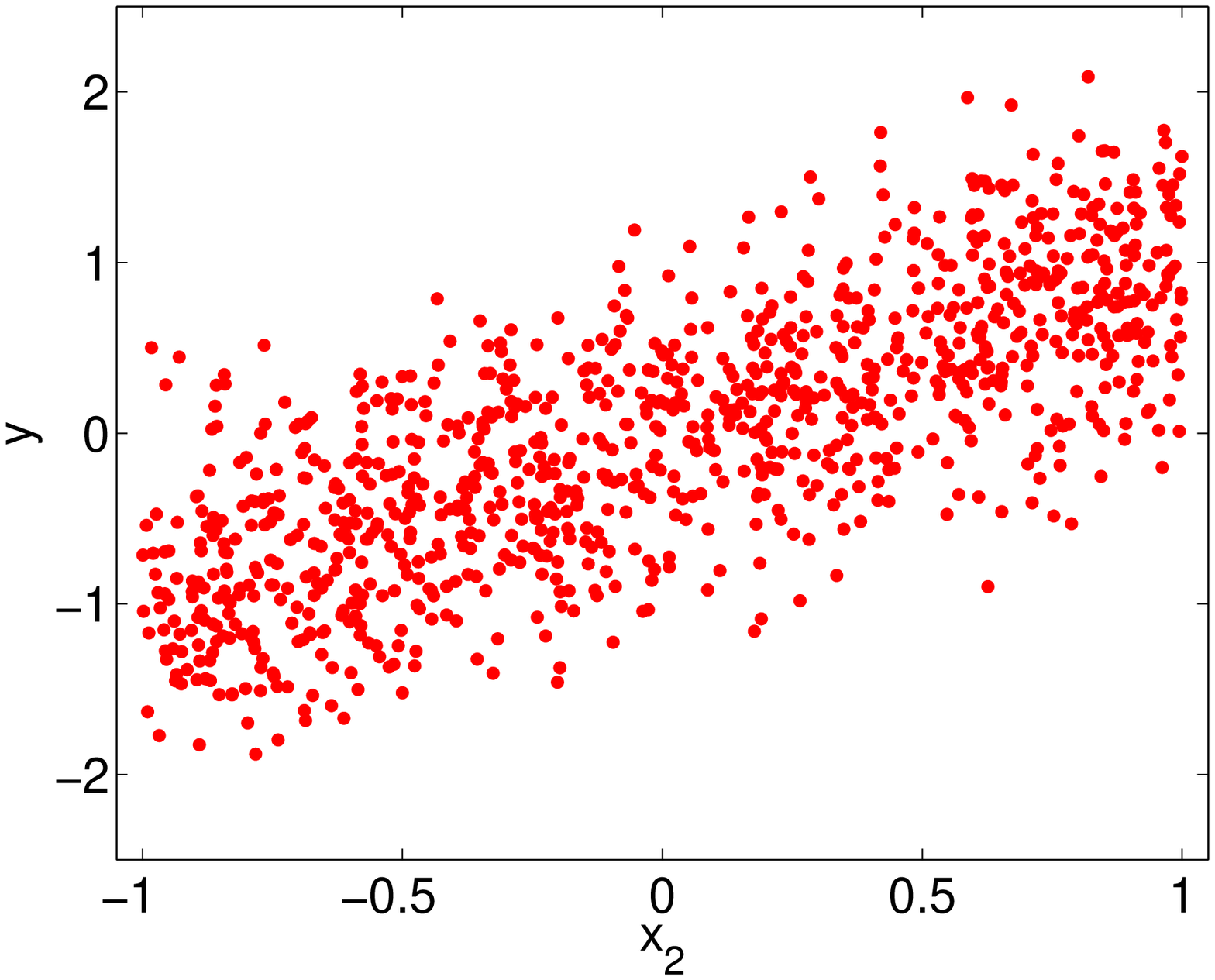}
\label{fig:senv_1}
}
\subfigure{
\includegraphics[width=.23\textwidth]{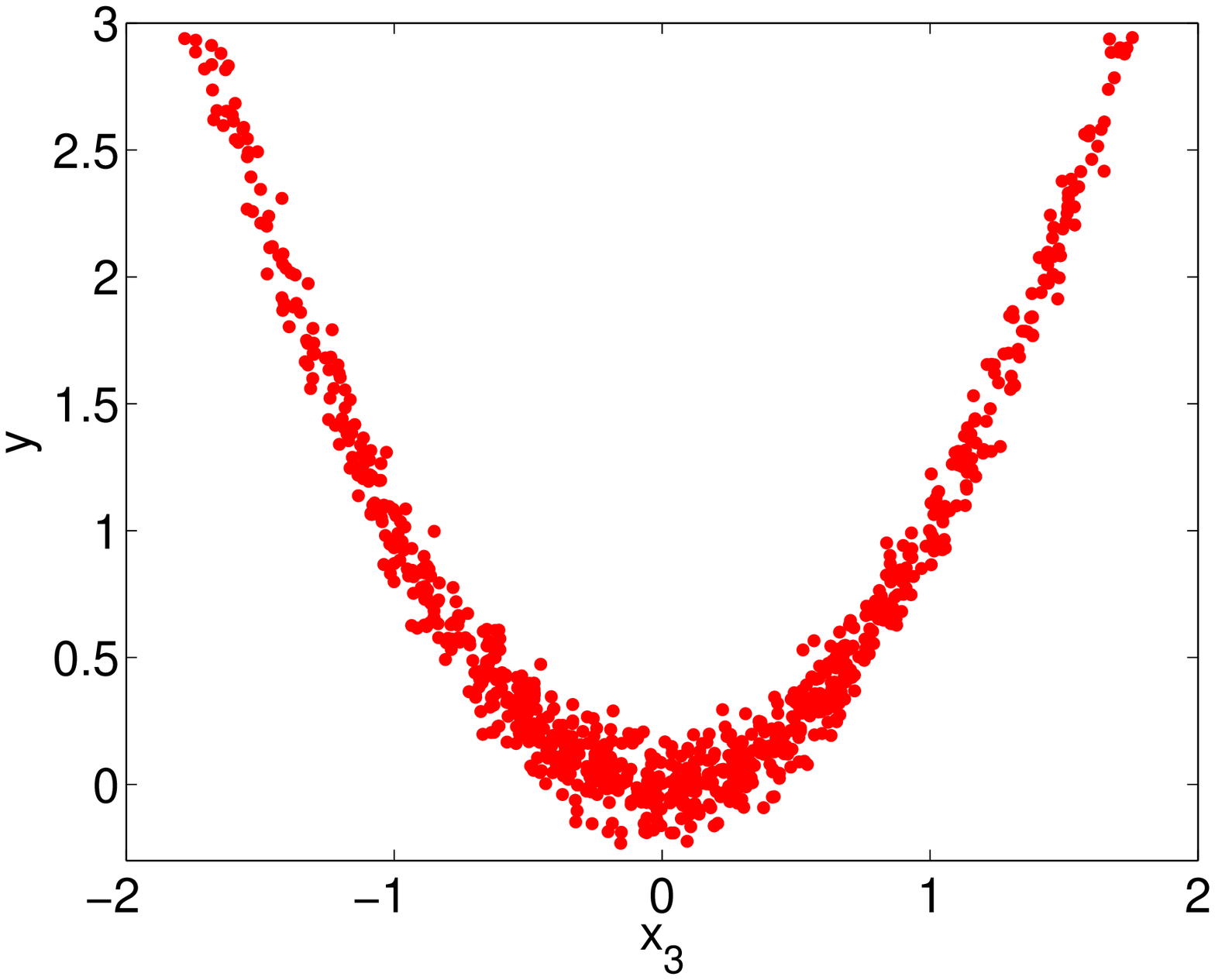}
  \label{fig:senv_2}
}
\subfigure{
\includegraphics[width=.23\textwidth]{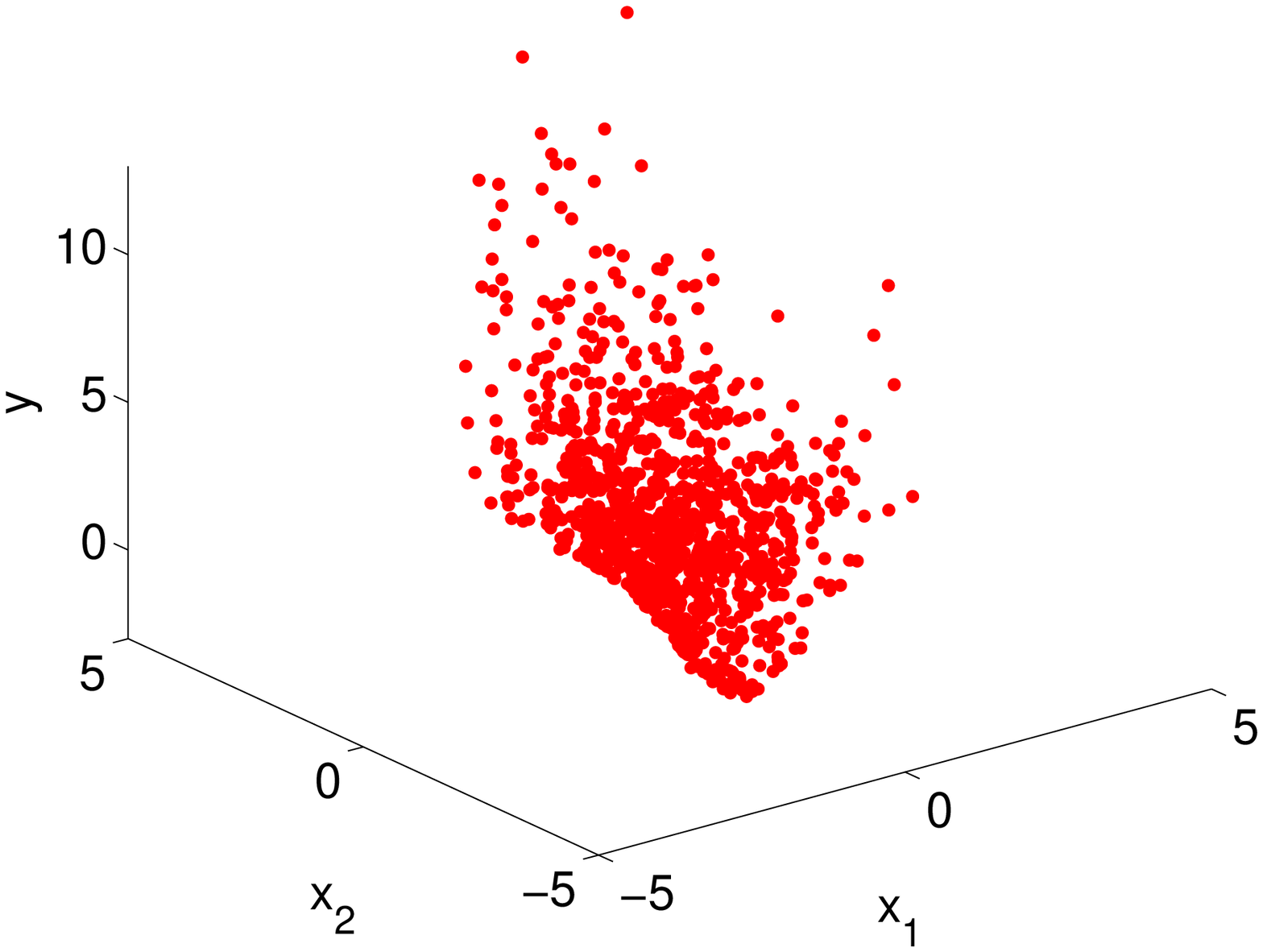}
  \label{fig:senv_3}
}
\subfigure{
\includegraphics[width=.23\textwidth]{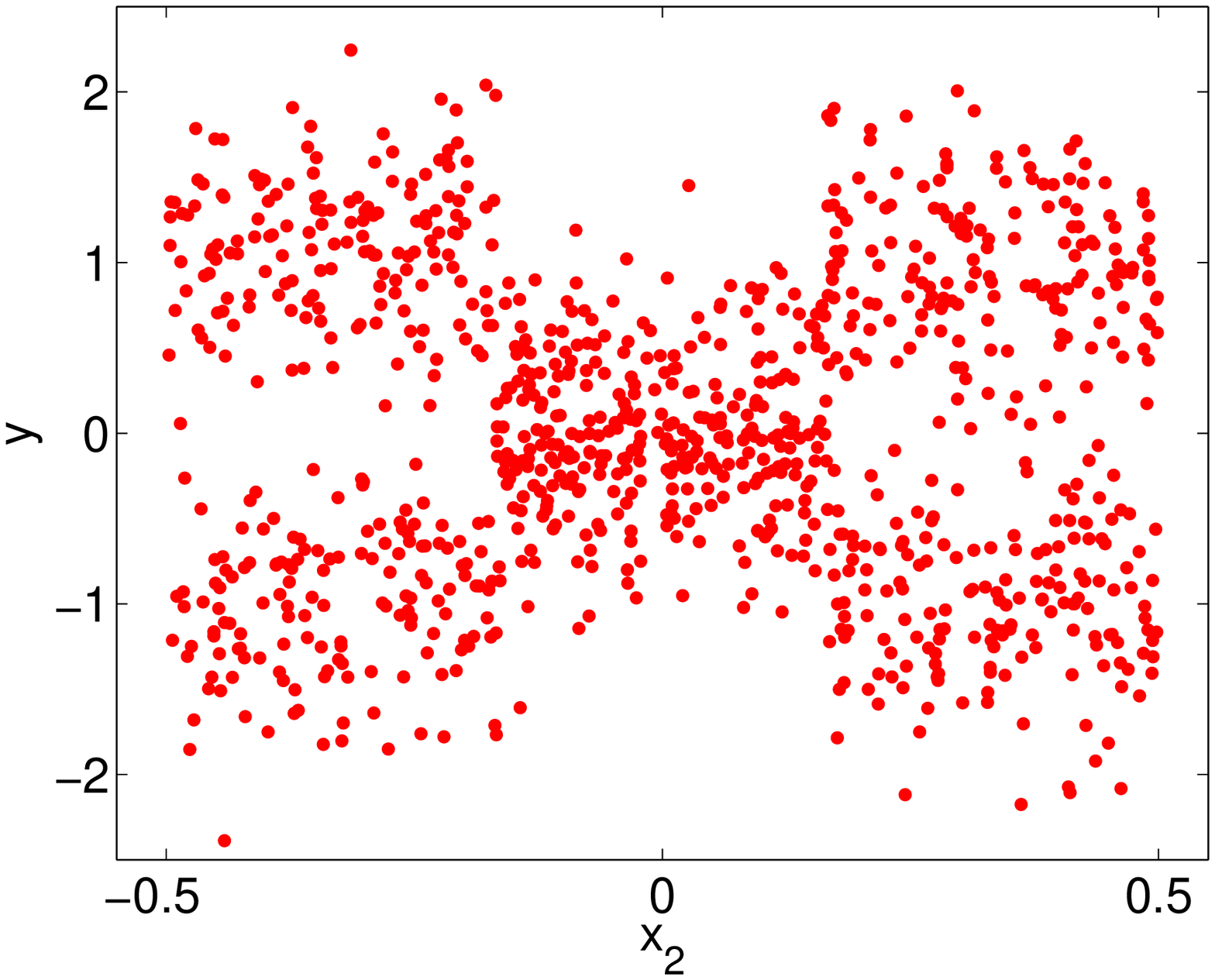}
  \label{fig:senv_3}
}
 \caption{Artificial datasets.
 }
    \label{fig:artificial_data}
\end{figure*}

\begin{table*}[t]
{\small
\centering \caption{Mean of Frobenius-norm error (with standard deviations in
brackets) and mean CPU time over 100 trials.
Computation time is normalized so that LSDR is one. LSDR was repeated 5 times with random initialization and the transformation matrix with the minimum CV score was chosen as the final solution.
`SCA(0)' indicates the performance of the initial transformation matrix obtained by
the method described in Section~\ref{subsec:initialization}.
 The best method in terms of the mean Frobenius-norm and comparable methods according to the \emph{t-test} at the significance level $1\%$ are specified by bold face. }
\vspace*{2mm}
\begin{tabular}{cr@{~~}r|l@{}r|l@{}r|l@{}r|l@{}r|l@{}r|l@{}r|l@{}r}
\hline
Datasets  & $d$ & $m$ & \multicolumn{2}{c|}{SCA(0)} & \multicolumn{2}{c|}{SCA} & \multicolumn{2}{c|}{$\LSDRG$}  & \multicolumn{2}{c|}{KDR}  & \multicolumn{2}{c|}{SIR}& \multicolumn{2}{c|}{SAVE}& \multicolumn{2}{c}{pHd}\\
\hline
Data1 & 4 & 1 & .089 & (.042) & {\bf .048} & {\bf (.031)} & {\bf .056}~ & {\bf (.021)}  & {\bf .048}~& {\bf (.019)}  & .257~ & (.168)& .339~ & (.218) & .593~ & (.210) \\
Data2 & 10 & 1 &  .078 & (.019) &{\bf .007} & {\bf (.002)} & .039 &(.023)  & .024 &(.007)  & .431 & (.281) & .348 &(.206) & .443 & (.222) \\
Data3 & 4 & 2& .065 & (.035) &{\bf .018} & {\bf (.010)} & .090 & (.069) & {\bf .029} & {\bf (.119)} &  .362 & (.182) & .343 & (.213) & .437 &(.231)\\
Data4 & 5 & 1 & .118 & (.046)&{\bf .042} & {\bf (.030)} & .151 & (.296)  & .118 &(.238)  & .421 &(.268) & .356 & (.197) & .591 &(.205) \\
\hline
\multicolumn{3}{c|}{Time} &\multicolumn{2}{c|}{0.03}& \multicolumn{2}{c|}{0.49} &\multicolumn{2}{c|}{1.0} &  \multicolumn{2}{c|}{0.96}  &     \multicolumn{2}{c|}{$<$0.01}  &     \multicolumn{2}{c|}{$<$0.01} &     \multicolumn{2}{c}{$<$0.01} \\
\hline
\label{table:toy}
\end{tabular}}
\end{table*}

\section{Experiments}
In this section, we experimentally investigate the performance of the proposed and existing SDR methods using artificial and real-world datasets. 

\subsection{Artificial Datasets}
We use four artificial datasets, and compare 
the proposed SCA, LSDR\footnote[1]{{\tiny \url{http://sugiyama-www.cs.titech.ac.jp/~sugi/software/LSDR/index.html}}} \cite{AISTATS:Suzuki+Sugiyama:2010},
KDR\footnote[2]{We used the program code provided by one of the authors of
\citet{TAS:Fukumizu+etal:2009}, which `anneals' the Gaussian kernel width
over gradient iterations.
} \cite{TAS:Fukumizu+etal:2009},
sliced inverse regression (SIR)\footnote[3]{{\tiny \url{http://mirrors.dotsrc.org/cran/web/packages/dr/index.html}}} \cite{JASA:Li:1991},
 sliced average variance estimation (SAVE)$^3$ \cite{TAM:Cook:2000}, and principal Hessian direction (pHd)$^3$ \cite{JASA:Li:1992}.

In SCA, we use the Gaussian kernel for $\boldy$:
\begin{align*}
L(\boldy,\boldy_\ell) = \exp \left(-\frac{\| \boldy - \boldy_\ell \|^2}{2\sigma_{\mathrm y}} \right).
\end{align*}
The identity matrix is used as regularization matrix $\boldR$,
and the kernel widths $\sigma_\mathrm{x}$, $\sigma_\mathrm{y}$, and $\sigma_\mathrm{z}$
as well as the regularization parameter $\lambda$ are
chosen based on 5-fold cross-validation.


The performance of each method is measured by
\[
\frac{1}{\sqrt{2m}} \|\widehat{\boldW}^\top \widehat{\boldW} - \boldW^{\ast \top}\boldW^\ast\|_{\mathrm{Frobenius}},
\]
where $\|\cdot\|_{\mathrm{Frobenius}}$ denotes the Frobenius norm, $\widehat{\boldW}$ is an estimated transformation matrix, and $\boldW^\ast$ is the optimal transformation matrix. Note that the above error measure takes its value in $[0,1]$.

We use the following four datasets (see Figure~\ref{fig:artificial_data}):
\begin{description}
\item[(a) Data1:] 
\[
Y = X_2 + 0.5E,
\] 
where $(X_1,\ldots,X_4)^\top \sim U([-1~1]^4)$ and $E \sim N(0,1)$.
Here, $U(\calS)$ denotes the uniform distribution on $\calS$,
and $N(\boldmu,\boldSigma)$ is the Gaussian distribution
with mean $\boldmu$ and variance $\boldSigma$.
\item[(b) Data2:] 
\[
Y = (X_3)^2 + 0.1E,
\] 
where $(X_1,\ldots,X_{10})^\top \sim N(\boldzero_{10}, \boldI_{10})$ and $E \sim N(0,1)$.
\item[(c) Data3:] 
\[
Y = \frac{(X_1)^2 + X_2}{0.5 + (X_2 + 1.5)^2} + (1+X_2)^2 + 0.1E,
\] 
where $(X_1,\ldots,X_4)^\top \sim N(\boldzero_4, \boldI_4)$ and $E \sim N(0,1)$.
\item[(d) Data4:] 
\begin{eqnarray*}
Y|X_2 \sim \left\{ \begin{array}{ll}
N(0, 0.2) & \textnormal{if}~X_2 \leq |1/6| \\
0.5N(1,0.2)  & \textnormal{otherwise} \\
+ 0.5N(-1,0.2), \\
\end{array} \right.
\end{eqnarray*} 
where $(X_1,\ldots,X_5)^\top \sim U([-0.5~0.5]^5)$ and $E \sim N(0,1)$.
\end{description}

The performance of each method is summarized in Table~\ref{table:toy}, which depicts the mean and standard deviation of the Frobenius-norm error over 100 trials when the number of samples is $n = 1000$. As can be observed, the proposed SCA overall performs well.
`SCA(0)' in the table indicates the performance of the initial transformation matrix obtained by
the method described in Section~\ref{subsec:initialization}.
The result shows that SCA(0) gives a reasonably good transformation matrix
with a tiny computational cost. Note that KDR and $\LSDRG$ have high standard deviation 
for Data3 and Data4, meaning that KDR and $\LSDRG$ sometimes perform poorly.

\subsection{Multi-label Classification for Real-world Datasets}
Finally, we evaluate the performance of the proposed method
in real-world multi-label classification problems.

\subsubsection{Setup}
Below, we compare SCA, Multi-label Dimensionality reduction via Dependence Maximization (MDDM)\footnote[4]{\tiny \url{http://cs.nju.edu.cn/zhouzh/zhouzh.files/publication/annex/MDDM.htm}} \cite{Zhang:2010:MDR:1839490.1839495}, Canonical Correlation Analysis (CCA)\footnote[5]{\tiny \url{http://www.mathworks.com/help/toolbox/stats/canoncorr.html}} \cite{Biometrika:Hotelling:1936}, and Principal Component Analysis (PCA)\footnote[6]{\tiny \url{http://www.mathworks.com/help/toolbox/stats/princomp.html}} \cite{book:Bishop:2006}.
We use a real-world image classification dataset called
the \emph{PASCAL Visual Object Classes (VOC) 2010} dataset \cite{pascal-voc-2010}
and a real-world automatic audio-tagging dataset called
the \emph{Freesound} dataset \cite{ref:freesound}.
Since the computational costs of KDR and LSDR were unbearably large,
we decided not to include them in the comparison.

We employ the misclassification rate by the nearest-neighbor classifier as a performance measure:
\[
 \mathrm{err} = \frac{1}{nc} \sum_{i = 1}^n \sum_{k = 1}^c I(\widehat{y}_{i,k} \neq y_{i,k}),
\]
where $c$ is the number of classes,
$\widehat{y}$ and $y$ are the estimated and true labels, and $I(y \neq y')$ is the indicator function.

For SCA and MDDM, we use the following kernel function  \cite{Sarwar:2001:ICF:371920.372071} for $\boldy$:
\begin{align*}
L(\boldy,\boldy') = \frac{(\boldy -\overline{\boldy})^\top
(\boldy' - \overline{\boldy})}{\| \boldy - \overline{\boldy}\|
 \| \boldy' - \overline{\boldy}'\|},
\end{align*}
where $\overline{\boldy}$ is the sample mean:
$\overline{\boldy}=\frac{1}{n}\sum_{i = 1}^n \boldy_i$.

\subsubsection{PASCAL VOC 2010 Dataset}
The VOC 2010 dataset consists of 20 binary classification tasks of
identifying the existence of a person, aeroplane, etc.~in each
image. The total number of images in the dataset is 11319, and we
used 1000 randomly chosen images for training and the rest for
testing.

In this experiment, we first extracted visual features from each image using the
\emph{Speed Up Robust Features} (SURF) algorithm
\cite{CVIU:Bay+etal:2008}, and obtained 500 \emph{visual words} as
the cluster centers in the SURF space. 
Then, we computed a 500-dimensional \emph{bag-of-feature} vector by counting the number of visual words
in each image. We randomly sampled the training and test data 100 times, and computed the means and standard deviations of the classification error.

\begin{figure*}[t]
  \centering
\subfigure[VOC 2010 dataset]{
\includegraphics[width=.45\textwidth]{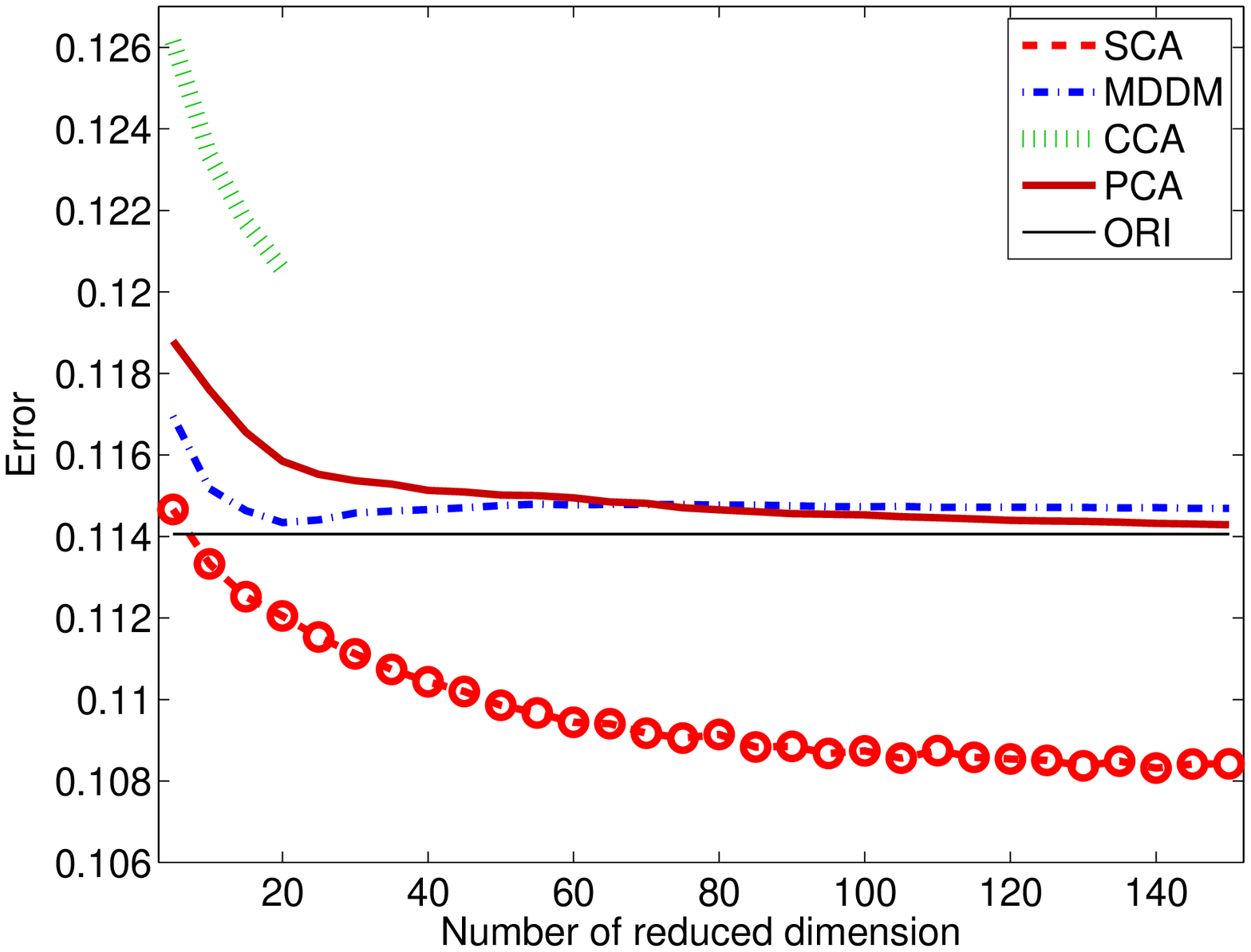}
  \label{fig:voc_result}
}
\subfigure[Freesound dataset]{
\includegraphics[width=.45\textwidth]{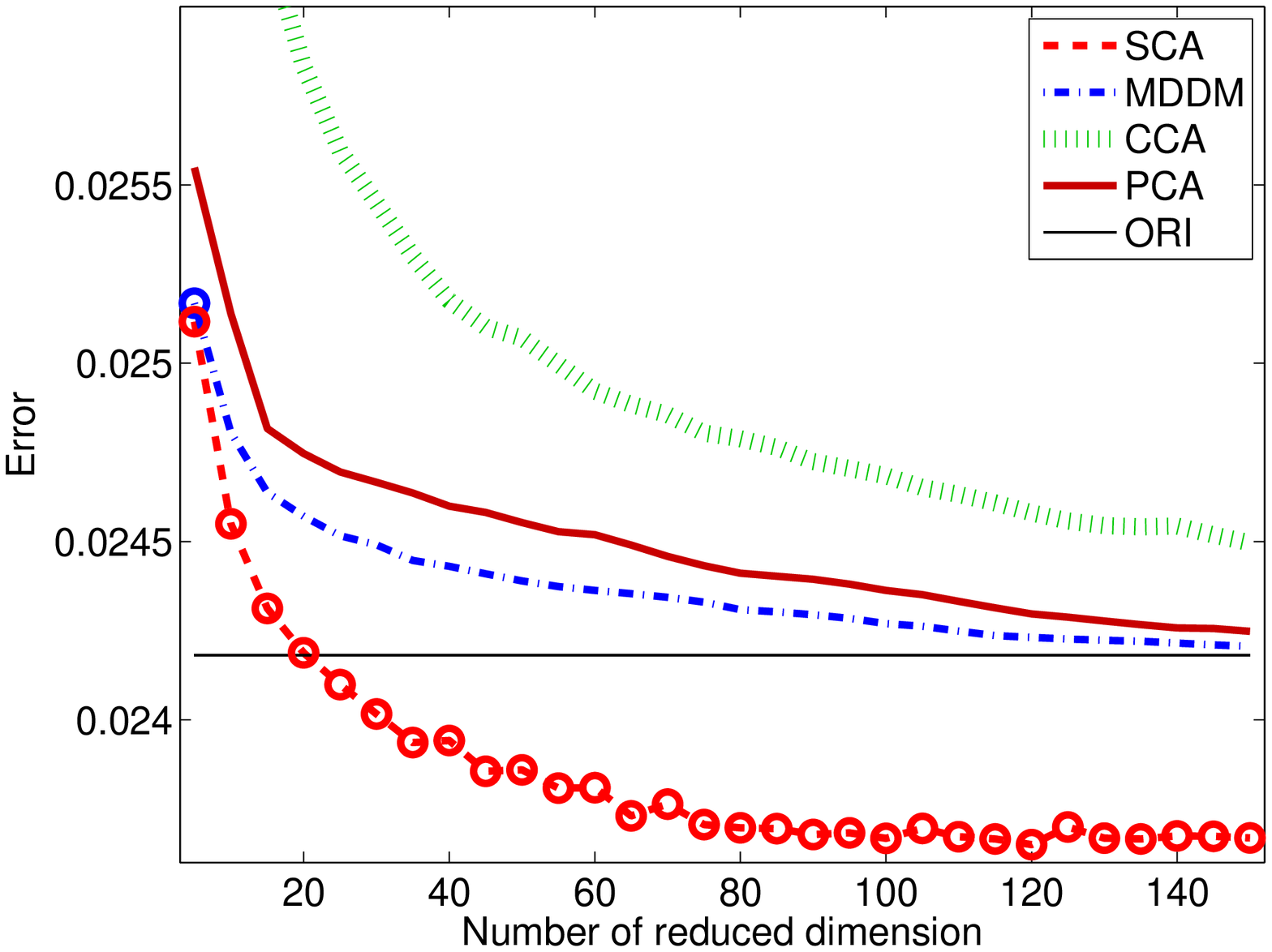}
  \label{fig:freesound_result}
}
 \caption{Results on image classification with VOC 2010 dataset and audio classification with Freesound datasets. Misclassification rate when the one-nearest-neighbor classifier is used as a classifier is reported. The best dimension reduction method in terms of the mean error and comparable methods according to the t-test at the significance level 1\% are specified by `$\circ$'. 
CCA can be applied to dimension reduction up to $c$ dimensions, where $c$ is the number of classes
($c=20$ in VOC 2010 and $c=230$ in Freesound).
`ORI' denotes the original data without dimension reduction.
}
    \label{fig:voc_freesound_result}
\end{figure*}

The results are plotted in Figure~\ref{fig:voc_result},
showing that SCA outperforms the existing methods,
and SCA is the only method that outperforms `ORI' (no dimension reduction)---SCA
achieves almost the same error rate as `ORI' with only a 10-dimensional subspace.

\subsubsection{Freesound Dataset}
\label{sec:audio-tagging} The \emph{Freesound} dataset
\cite{ref:freesound} consists of various audio files annotated with
word tags such as `people', `noisy', and `restaurant'.
We used 230 tags in this experiment. The total number of audio files in the dataset is 5905, and we used 1000 randomly chosen audio files for training and the rest for testing.

We first extracted \emph{Mel-Frequency Cepstrum Coefficients} (MFCC) \cite{book:Rabiner:1993} from each audio file, and obtained 1024 \emph{audio features} as
the cluster centers in MFCC. 
 Then, we computed a 1024-dimensional
\emph{bag-of-feature} vector by counting the number of audio features
in each audio file. We randomly chose the training and test samples
100 times, and computed the means and standard deviations of the classification error.

The results plotted in Figure~\ref{fig:freesound_result} show that,
similarly to the image classification task, the proposed SCA outperforms the existing methods,
and SCA is the only method that outperforms `ORI'.

\section{Conclusion}
In this paper, we proposed a novel \emph{sufficient dimension reduction} (SDR) method called \emph{sufficient component analysis} (SCA), which is computationally more efficient than existing SDR methods. In SCA, a transformation matrix was  estimated by iteratively performing dependence estimation and maximization, both of which are \emph{analytically} carried out.
Moreover, we developed a systematic method to design a good initial transformation matrix,
which highly contributes to further reducing the computational cost
and help obtain a good local optimum solution.
We applied the proposed SCA to real-world 
image classification  and audio tagging tasks,
and experimentally showed that the proposed method is promising.

\section*{Acknowledgments}
The authors thank Prof.~Kenji Fukumizu for providing us the KDR code and Prof.~Taiji Suzuki for his valuable comments.
MY was supported by the JST PRESTO program. GN was supported by the MEXT scholarship. MS was supported by
SCAT, AOARD, and the JST PRESTO program.


\bibliography{icml2011}

\begin{thebibliography}{18}
\providecommand{\natexlab}[1]{#1}
\providecommand{\url}[1]{\texttt{#1}}
\expandafter\ifx\csname urlstyle\endcsname\relax
  \providecommand{\doi}[1]{doi: #1}\else
  \providecommand{\doi}{doi: \begingroup \urlstyle{rm}\Url}\fi

\bibitem[Amari(1998)]{NECO:Amari:1998}
Amari, S.
\newblock Natural gradient works efficiently in learning.
\newblock \emph{Neural Computation}, 10:\penalty0 251--276, 1998.

\bibitem[Bay et~al.(2008)Bay, Ess, Tuytelaars, and Gool]{CVIU:Bay+etal:2008}
Bay, H., Ess, A., Tuytelaars, T., and Gool, L.~V.
\newblock Surf: Speeded up robust features.
\newblock \emph{Computer Vision and Image Understanding}, 110\penalty0
  (3):\penalty0 346--359, 2008.

\bibitem[Bishop(2006)]{book:Bishop:2006}
Bishop, C.~M.
\newblock \emph{Pattern Recognition and Machine Learning}.
\newblock Springer, New York, NY, 2006.

\bibitem[Cook(1998)]{book:Cook:1998}
Cook, R.~D.
\newblock \emph{Regression graphics: Ideas for studying regressions through
  graphics}.
\newblock Wiley, New York, 1998.

\bibitem[Cook(2000)]{TAM:Cook:2000}
Cook, R.~D.
\newblock Save: A method for dimension reduction and graphics in regression.
\newblock \emph{Theory and Methods}, 29:\penalty0 2109--2121, 2000.

\bibitem[Epanechnikov(1969)]{TPA:Epanechnikov:1969}
Epanechnikov, V.
\newblock Nonparametric estimates of a multivariate probability density.
\newblock \emph{Theory of Probability and its Applications}, 14:\penalty0
  153--158, 1969.

\bibitem[Everingham et~al.(2010)Everingham, Gool, Williams, Winn, and
  Zisserman]{pascal-voc-2010}
Everingham, M., Gool, L.~V., Williams, C. K.~I., Winn, J., and Zisserman, A.
\newblock The {PASCAL} {V}isual {O}bject {C}lasses {C}hallenge 2010 {(VOC2010)}
  {R}esults.
\newblock http://www.pascal-network.org/challenges/VOC/voc2010/workshop/
  index.html, 2010.

\bibitem[Fukumizu et~al.(2009)Fukumizu, Bach, and
  Jordan]{TAS:Fukumizu+etal:2009}
Fukumizu, K., Bach, F.~R., and Jordan, M.
\newblock Kernel dimension reduction in regression.
\newblock \emph{The Annals of Statistics}, 37\penalty0 (4):\penalty0
  1871--1905, 2009.

\bibitem[Hotelling(1936)]{Biometrika:Hotelling:1936}
Hotelling, H.
\newblock Relations between two sets of variates.
\newblock \emph{Biometrika}, 28:\penalty0 321--377, 1936.

\bibitem[Li(1991)]{JASA:Li:1991}
Li, K.-C.
\newblock Sliced inverse regression for dimension reduction.
\newblock \emph{Journal of American Statistical Association}, 86:\penalty0
  316--342, 1991.

\bibitem[Li(1992)]{JASA:Li:1992}
Li, K.-C.
\newblock On principal {H}essian directions for data visualization and
  dimension reduction: Another application of {S}teinfs lemma.
\newblock \emph{Journal of American Statistical Association}, 87:\penalty0
  1025--1034, 1992.

\bibitem[Patriksson(1999)]{Book:Patriksson:1999}
Patriksson, M.
\newblock \emph{Nonlinear Programming and Variational Inequality Problems}.
\newblock Kluwer Academic, Dredrecht, 1999.

\bibitem[Rabiner \& Juang(1993)Rabiner and Juang]{book:Rabiner:1993}
Rabiner, L. and Juang, B-H.
\newblock \emph{Fundamentals of Speech Recognition}.
\newblock Prentice Hall, Englewood Cliffs, NJ, 1993.

\bibitem[Sarwar et~al.(2001)Sarwar, Karypis, Konstan, and
  Reidl]{Sarwar:2001:ICF:371920.372071}
Sarwar, B., Karypis, G., Konstan, J., and Reidl, J.
\newblock Item-based collaborative filtering recommendation algorithms.
\newblock In \emph{Proceedings of the 10th international conference on World
  Wide Web (WWW2001)}, pp.\  285--295, 2001.

\bibitem[Suzuki \& Sugiyama(2010)Suzuki and
  Sugiyama]{AISTATS:Suzuki+Sugiyama:2010}
Suzuki, T. and Sugiyama, M.
\newblock Sufficient dimension reduction via squared-loss mutual information
  estimation.
\newblock In \emph{Proceedings of the Thirteenth International Conference on
  Artificial Intelligence and Statistics (AISTATS2010)}, pp.\  804--811, 2010.

\bibitem[Suzuki et~al.(2009)Suzuki, Sugiyama, Kanamori, and
  Sese]{BMCBio:Suzuki+etal:2009a}
Suzuki, T., Sugiyama, M., Kanamori, T., and Sese, J.
\newblock Mutual information estimation reveals global associations between
  stimuli and biological processes.
\newblock \emph{BMC Bioinformatics}, 10\penalty0 (S52), 2009.

\bibitem[{The Freesound Project}(2011)]{ref:freesound}
{The Freesound Project}.
\newblock {Freesound}, 2011.
\newblock http://www.freesound.org.

\bibitem[Zhang \& Zhou(2010)Zhang and Zhou]{Zhang:2010:MDR:1839490.1839495}
Zhang, Y. and Zhou, Z.-H.
\newblock Multilabel dimensionality reduction via dependence maximization.
\newblock \emph{ACM Trans. Knowl. Discov. Data}, 4:\penalty0 14:1--14:21, 2010.
\newblock ISSN 1556-4681.

\end{thebibliography}
\bibliographystyle{icml2011}

\end{document}